\documentclass{ecai} 
\usepackage{latexsym}
\usepackage{amssymb}
\usepackage{amsmath}
\usepackage{amsthm}
\usepackage{booktabs}
\usepackage{enumitem}
\usepackage{graphicx}
\usepackage{color}

\usepackage{hyperref}
\usepackage{url}
\usepackage{xcolor}
\usepackage{subfig}
\usepackage{booktabs}

\newcommand{\nummodels}{144}

\DeclareMathOperator*{\argmax}{arg\,max}

\newcommand{\BibTeX}{B\kern-.05em{\sc i\kern-.025em b}\kern-.08em\TeX}

\begin{document}

%%%%%%%%%%%%%%%%%%%%%%%%%%%%%%%%%%%%%%%%%%%%%%%%%%%%%%%%%%%%%%%%%%%%%%%%

\begin{frontmatter}

%%% Use this command to specify your submission number.
%%% In doubleblind mode, it will be printed on the first page.

\paperid{5395} 

%%% Use this command to specify the title of your paper.

\title{Alignment and Adversarial Robustness: Are More Human-Like Models More Secure?}

%%% Use this combinations of commands to specify all authors of your 
%%% paper. Use \fnms{} and \snm{} to indicate everyone's first names 
%%% and surname. This will help the publisher with indexing the 
%%% proceedings. Please use a reasonable approximation in case your 
%%% name does not neatly split into "first names" and "surname".
%%% Specifying your ORCID digital identifier is optional. 
%%% Use the \thanks{} command to indicate one or more corresponding 
%%% authors and their email address(es). If so desired, you can specify
%%% author contributions using the \footnote{} command.

\author[A]{\fnms{Blaine}~\snm{Hoak}\orcid{0000-0003-2960-0686}\thanks{Corresponding Author. Email: bhoak@cs.wisc.edu}\footnote{Equal contribution.}}
\author[A]{\fnms{Kunyang}~\snm{Li}\orcid{0009-0007-4264-855X}\footnotemark}
\author[A]{\fnms{Patrick}~\snm{McDaniel}\orcid{0000-0003-2091-7484}} 

\address[A]{University of Wisconsin-Madison}

%%% Use this environment to include an abstract of your paper.

\begin{abstract}
% Representational alignment refers to the extent to which a model’s internal representations mirror biological vision, offering insights into both neural similarity and functional correspondence. Recently, some more aligned models have demonstrated higher resiliency to adversarial examples, raising the question of whether more human-aligned models are inherently more secure. 
A small but growing body of work has shown that machine learning models which better align with human vision have also exhibited higher robustness to adversarial examples, raising the question: \textit{can human-like perception make models more secure?} If true generally, such mechanisms would offer new avenues toward robustness.
% Recent work has shown that some machine learning models which better align with human vision are also more robust to adversarial examples, raising the question: \textit{can human-like perception make models more secure?} 
In this work, we conduct a large-scale empirical analysis to systematically investigate the relationship between representational alignment and adversarial robustness. We evaluate \nummodels{} models spanning diverse architectures and training paradigms, measuring their neural and behavioral alignment and engineering task performance across 105 benchmarks as well as their adversarial robustness via AutoAttack. Our findings reveal that while average alignment and robustness exhibit a weak overall correlation, \textit{specific} alignment benchmarks serve as strong predictors of adversarial robustness, particularly those that measure selectivity toward texture or shape. These results suggest that different forms of alignment play distinct roles in model robustness, motivating further investigation into how alignment-driven approaches can be leveraged to build more secure and perceptually-grounded vision models.
\end{abstract}

\end{frontmatter}

%%%%%%%%%%%%%%%%%%%%%%%%%%%%%%%%%%%%%%%%%%%%%%%%%%%%%%%%%%%%%%%%%%%%%%%%

\section{Introduction}\label{sec:intro}
% A longstanding goal in computer vision is to develop models that process images in a way that aligns with human perception. 
Representational alignment---how closely a model resembles biological vision---has been studied extensively with the goal of measuring, bridging, or increasing alignment in machine learning models~\cite{sucholutsky_getting_2024}. 
Recent observations~\cite{dapello_simulating_2020,li_learning_2019} suggest that alignment may have implications beyond neuroscience: models that are more aligned with human perception have also exhibited increased robustness to adversarial examples---inputs with near-imperceptible perturbations that induce model misclassification--- hinting at a deeper connection between alignment and security.

However, the relationship between representational alignment and adversarial robustness remains poorly understood. While the former seeks to align models with human cognition, adversarial examples in security highlight a fundamental misalignment: imperceptible perturbations can drastically degrade model accuracy while leaving human perception unaffected. Prior robustness techniques, such as adversarial training~\cite{madry_towards_2019}, are computationally expensive and potentially vulnerable to new attack strategies. 
% Meanwhile, alignment research has not systematically examined whether more human-aligned models are inherently more robust to adversarial attacks. 
Although recent work has suggested links between perceptual alignment and robustness, especially in models incorporating biological priors~\cite{dapello_simulating_2020, li_learning_2019}, a broad and systematic evaluation of this relationship across diverse models and alignment benchmarks is still lacking.
A fundamental question remains: do these objectives complement each other, leading to better-aligned and more robust models, or do they introduce conflicting trade-offs?

% Solution
In this work, we investigate the relationship between human alignment and robustness to adversarial examples in vision models through a diverse, large-scale empirical analysis. In our analysis, we study \nummodels{} models across different architectures and training schemes, and measure their alignment across 105 different benchmarks on neural, behavioral, and engineering tasks via the BrainScore library~\cite{schrimpf_brain-score_2018}. We then evaluate the robustness of these models using AutoAttack~\cite{croce_reliable_2020}, a state-of-the-art ensemble attack. 

% In analyzing the correlations between model robustness and alignment, our findings reveal that while robustness is weakly correlated with vision alignment on average, certain alignment benchmarks serve as strong indicators of model robustness. Specifically, we find that the top six benchmarks that were most positively correlated with robust accuracy, even with strong perturbations, \textit{all measured a model's selectivity towards texture information}. 

% In addition, our t-SNE analysis of \nummodels{} models across 105 benchmarks shows that models with similar performance on alignment benchmarks tend to exhibit similar degree of adversarial robustness. 
% % This demonstrates that alignment benchmark similarity provides partial predictive power for robustness, suggesting that standard alignment metrics do not fully capture the underlying factors determining adversarial robustness. 
% Overall, these results imply that different forms of alignment play distinct roles in model robustness, motivating further investigation into how alignment-driven approaches can be leveraged to build more secure and perceptually-grounded vision models.

Our analysis reveals that while robustness is only weakly correlated with \textit{average} vision alignment, \textit{specific} alignment benchmarks are strongly predictive of adversarial robustness. We also find that models with similar alignment profiles exhibit similar robustness. Together, these findings suggest that different forms of alignment contribute differently to robustness, highlighting the value of alignment-driven approaches for improving security in vision systems.

\begin{itemize}
    \item \textit{Average} alignment, particularly for behavioral benchmarks, poorly predicts robustness (explaining 6\% of variance), demonstrating that not all methods of alignment increase robustness.
    \item Individual benchmarks are much better indicators of robust accuracy, but the \textit{specific} benchmark matters; we found instances of benchmarks that contributed positively \textit{and} ones that contributed negatively to robustness in every alignment category.
    \item The top benchmarks that are most indicative of robust accuracy show clear trends that inform the factors leading to higher robustness: (1) robust models process \textit{texture information} specifically more similarly to humans than non-robust models to and (2) models' ability to recognize objects without global structures intact actually hurts their robustness.
    \item t-SNE visualizations reveal that robustness is structured in alignment space, with similar models forming robustness-consistent clusters and demonstrating that alignment on certain benchmarks is a good indicator of robustness. 
\end{itemize}

\noindent In summary, we uncover specific features of alignment that closely tie with model robustness and show that increasing alignment on these benchmarks offers a new avenue for building more robust and human-aligned models.

\section{Background}\label{sec:background}
\subsection{Representational Alignment}
Representational alignment studies the extent to which internal representations of machine learning models correspond to human cognitive processes. 
Early studies found that deep neural networks (DNNs) trained on large-scale image datasets develop hierarchical feature representations similar to those observed in the primate ventral stream, particularly in high-level visual areas like the inferior temporal (IT) cortex \cite{yamins_hierarchical_2013, schrimpf_brain-score_2018}. This led to efforts to quantify the alignment between artificial and biological vision, using techniques such as Representational Similarity Analysis (RSA) \cite{kriegeskorte_representational_2008} and Centered Kernel Alignment (CKA) \cite{kornblith_similarity_2019}. 
% Current research in the area primarily focuses on measuring, bridging, and increasing both neural and behavioral alignment. 
To improve alignment, researchers have proposed strategies that incorporate cognitive constraints or psychological priors into model architectures~\cite{dapello_simulating_2020}. Supervised fine-tuning with human-annotated datasets~\cite{dosovitskiy_image_2021} ensures that learned representations align more closely with human-understandable features. Furthermore, novel techniques~\cite{muttenthaler_improving_2023,li_learning_2019,cheng_rtify_2024} have been developed to encourage similarity between model activations and human neural responses as recorded through fMRI and EEG experiments. In this study, we use a comprehensive set of neural, behavioral, and engineering alignment metrics to quantify representational alignment.

\subsection{Adversarial Examples}
% Why do people care about adversarial examples? 
Although machine learning models have shown strong capabilities to achieve high accuracy in various tasks~\cite{liu_convnet_2022, dosovitskiy_image_2021, krizhevsky_imagenet_2017, he_deep_2015}, they remain vulnerable to adversarial examples~\cite{croce_reliable_2020,madry_towards_2019, carlini_towards_2017,goodfellow_explaining_2015, sheatsley_space_2023}. Adversarial examples are specially crafted inputs that contain perturbations which are imperceptible to humans, yet can significantly decrease model accuracy. In computer vision systems, there have been many studies on developing attack algorithms, such as FGSM~\cite{goodfellow_explaining_2015}, PGD~\cite{madry_towards_2019}, and AutoAttack~\cite{croce_reliable_2020}. These methods  aim to maximize model's loss subject to constraints of perturbations defined by certain $\ell_p$-norms:

\begin{center}
    $x_{adv} = \argmax_{\left \| \delta \right \|_{p}\leq\epsilon} L(x + \delta, y)$
\end{center}

\noindent where $x$ and $y$ represent the original image and its predicted label, respectively, $\delta$ is the perturbation to solve for, and $L$ is the model's loss function. The perturbation constraint $\epsilon$ is measured through an $\ell_p$-norm---most commonly $\ell_\infty$. While many works have historically evaluated the robustness of their model through  PGD~\cite{madry_towards_2019}, it has been shown that ``robust'' models can often suffer from gradient masking, causing gradient-based attacks like PGD to fail~\cite{athalye_obfuscated_2018}, and leading to a sense of overestimated robustness. To overcome this, multiple attacks, including both white- and black-box attacks should be used~\cite{carlini_evaluating_2019}. Thus, the  AutoAttack ensemble~\cite{croce_reliable_2020} has become the de-facto standard for evaluating robustness.

\section{Methods}\label{sec:methods}
This section details how we measure the alignment and robustness of machine learning models with the goal of assessing if more human-like vision models are also more resilient to security vulnerabilities.

\subsection{Alignment.} To measure alignment and download candidate models, we leverage the BrainScore~\cite{schrimpf_brain-score_2018} library. BrainScore provides a standardized framework for evaluating model similarity to biological vision through a set of neural, behavioral, and engineering benchmarks, supplying 106 benchmarks in total. These benchmarks quantify how closely a model’s internal representations and outputs correspond to neurophysiological recordings, human psychophysical behavior, and performance on engineered vision tasks. Neural alignment is measured by comparing activations from DNNs to neural recordings from primate visual cortex regions (e.g., V1, V2, V4, and IT), using similarity metrics like Representational Similarity Analysis (RSA) \cite{kriegeskorte_representational_2008}. Behavioral alignment assesses whether models replicate human psycho-physical responses in object recognition and perturbation tests, while engineering alignment evaluates model robustness to controlled distortions, such as contrast reductions, or performance on out-of-distribution data. We use 105 benchmarks by discarding one of them, which has NaN value for all the selected models.

In total, the BrainScore library has documented benchmark scores for 434 models. Out of those, there are 240 models available in their registry (the remaining models were either submitted privately or have been deprecated). From the 240 models in the registry, we further removed an additional 89 models that were either incompatible with ImageNet (e.g., do not output 1000 classes or expect video streams) or could not be run on RobustBench due to gradient masking. 
% because either loading the model produced a ClientError due to a moved or removed model hosting location or the model was incompatible with ImageNet (either does not output 1000 classes or expects video streams). 
After this, we had to discard an additional 7 models, which represented all the VOne class models~\cite{dapello_simulating_2020} because they were not able to run on AutoAttack due to gradient alteration or masking, suggesting that previous results finding that VOne models are more robust to adversarial examples could have been due to overestimated robustness and highlighting the importance of evaluating robustness under comprehensive attack strategies. After this filtering process, we were left with \nummodels{} models for our evaluation.

Model diversity is critical for evaluating the generalization of alignment-robustness relationships. Thus, the \nummodels{} models selected for our evaluation are representative by covering a broad spectrum of architectures and training recipes. The majority of architectures are convolutional neural networks (CNNs)~\cite{oshea_introduction_2015} such as various ResNet~\cite{he_deep_2015} and VGG variants~\cite{simonyan_deep_2014}. We also include more recent architectural designs such as Vision Transformers (ViTs)~\cite{dosovitskiy_image_2021}, which uses self-attention mechanisms to capture global dependencies in images, its variants~\cite{beyer_knowledge_2022}, and hybrid models~\cite{liu_convnet_2022,touvron_training_2021} (i.e., a combination of CNNs and ViTs). This architectural diversity is complemented by models trained with different strategies such as standard supervised learning and self-supervised learning (e.g., contrastive learning~\cite{radford_learning_2021}). Notably, the evaluated models also include those specifically designed or trained with properties relevant to the study's central hypothesis: some models have undergone adversarial training~\cite{madry_towards_2019}, which improves robustness against adversarial examples, while others use mechanisms to emphasize shape bias~\cite{geirhos_imagenet-trained_2019, geirhos_generalisation_2020, hermann_origins_2020}, shifting model reliance from texture to shape information. This design has been shown to be more aligned with human perception. The comprehensive model set allows for a thorough exploration of how architectural design and training methods impact the relationship between representational alignment and adversarial robustness.  

\subsection{Robustness.} To evaluate the robustness of our models, we use AutoAttack~\cite{croce_reliable_2020, croce_robustbench_2021}, which serves as the standard for evaluating the robustness of neural networks due to its strong attack performance and fully automated parameter-free design. AutoAttack contains 4 attacks: APGD-CE, APGD-DLR, FAB, and Square Attack. By evaluating on AutoAttack, we are not only evaluating on the most performant attacks, but also integrating in both white-box attacks and black-box attacks which has been recommended in previous works to combat reporting overestimated robustness due to gradient masking or obfuscation~\cite{carlini_towards_2017}.

To better understand how the relationship between adversarial robustness and alignment changes as attacks change, we evaluate the $\ell_\infty$ robustness of our models at three different epsilon levels: $\epsilon = \{\frac{0.25}{255}, \frac{0.5}{255}, \frac{1}{255}\}$ to represent adversaries at different capability levels and small, medium, and large image distortion levels. While these values are typically lower than what would be benchmarked on platforms such as RobustBench~\cite{croce_robustbench_2021}, we choose these values with the goal of having a wide distribution of robust accuracies to identify separability between models, rather than the goal of bringing the model down to 0\% accuracy as is typically done. 
\section{Results}\label{sec:results}
In this work, we hypothesize that there is a relationship between model robustness and alignment, due to the inherent similarity of the goals in each of these spaces. Here, we focus on answering the question \textit{are more aligned machine learning models more robust to adversarial examples?}

\begin{figure}[t!]
    \centering
    \subfloat[Neural Alignment]{\includegraphics[width=0.99\linewidth]{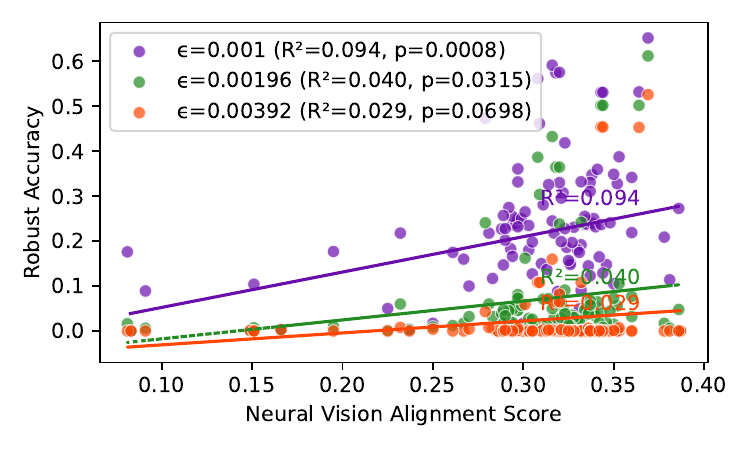}}
    \hfill 
    \subfloat[Behavioral Alignment]{\includegraphics[width=0.99\linewidth]{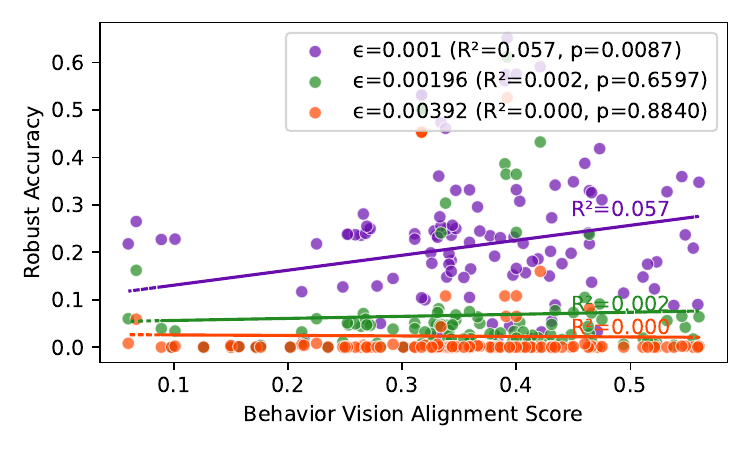}}
    \hfill 
    \subfloat[Engineering Task Performance]{\includegraphics[width=0.99\linewidth]{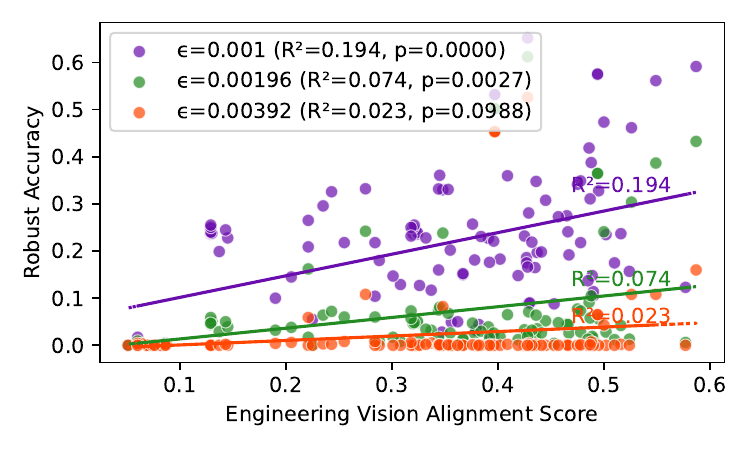}}
    \caption{Average vision alignment score vs robust accuracy on neural, behavioral, and engineering benchmarks.}
    \vspace{0.4cm}
    \label{fig:avg_vision_robustness}
    % \Description[Scatter plots of robust accuracy and BrainScore alignment]{Three scatter plots, measuring the neural alignment, behavioral alignment, and engineering task performance, respectively, versus robust accuracy. Each scatter plot has points of three different colors that separate the 3 different epsilon values used to generate the adversarial examples. A line for each color of point shows the R squared value and p value to measure the correlation between robust accuracy and vision alignment.}
    \vspace{0.4cm}
\end{figure}

\begin{figure*}[t!]
    \centering
    \includegraphics[width=\linewidth]{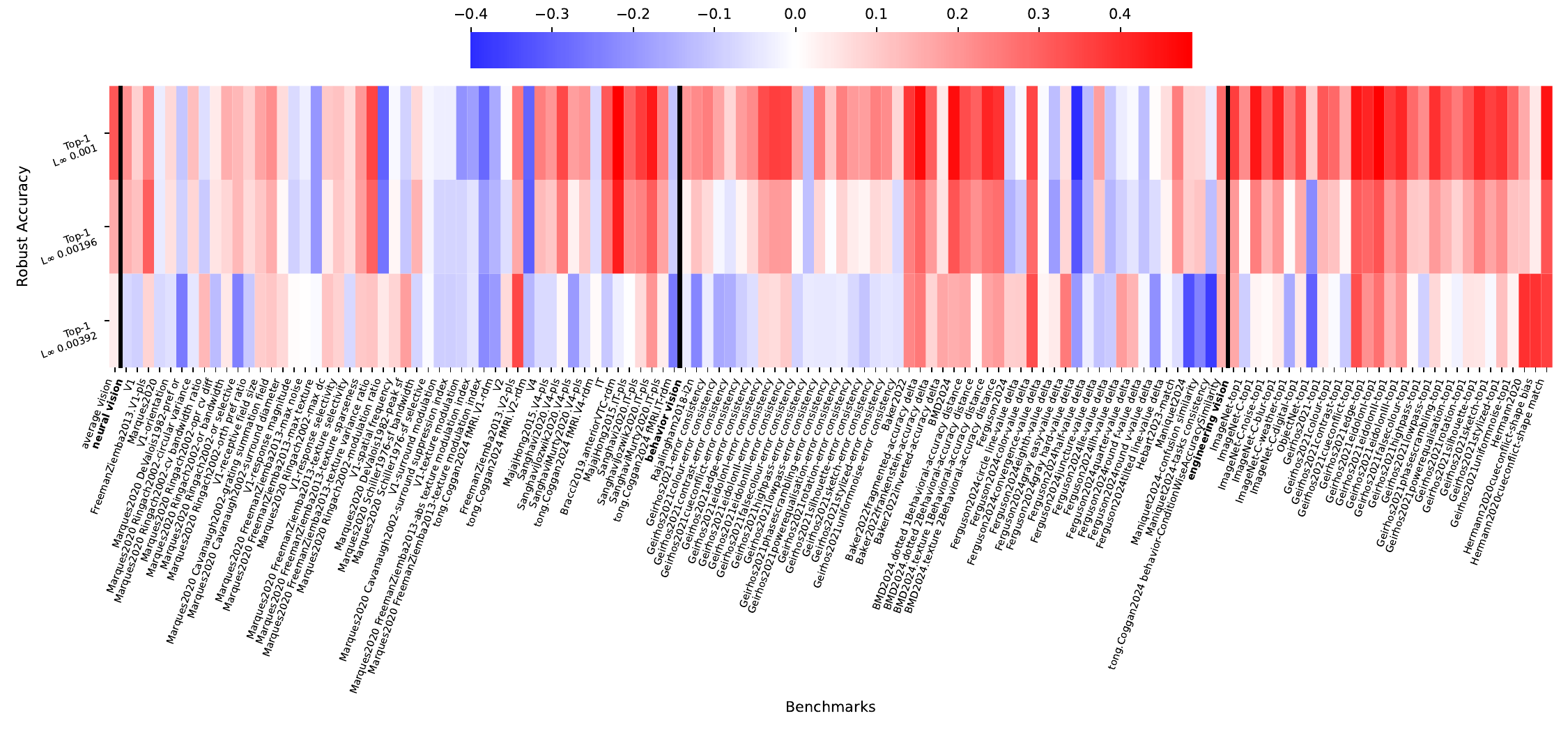}
    % \Description[A heatmap of BrainScore benchmarks and robust accuracy]{A heatmap of the correlation (across models) between BrainScore benchmarks on the x axis and robust accuracy on the y axis on L infinity adversarial examples at 3 different values, 0.001, 0.00196, and 0.00392. There are 3 solid black vertical lines that separate the 3 types of BrainScore benchmarks: neural vision, behavioral vision, and engineering vision.}
    \caption{Heatmap of each of the BrainScore benchmarks, ordered and separated (black bars) by area of alignment (neural, behavioral, engineering) vs the robust accuracy. Each cell represents the correlation between a benchmark and the robust accuracy across models.}
    \label{fig:robust_only_correlation}
    \vspace{0.3cm}
\end{figure*}

% \subsection{Experimental Details}
To address this, we conduct a series of experiments. We use the BrainScore library v2.2.4 to measure alignment~\cite{schrimpf_brain-score_2018} and load models. Details on models evaluated can be found in \autoref{sec:methods}. Once these models have been loaded and their alignment has been measured across the 105 alignment benchmarks, we evaluate their robustness using AutoAttack~\cite{croce_reliable_2020} from the TorchAttacks~\cite{kim_torchattacks_2021} library v3.5.1. The ImageNet~\cite{russakovsky_imagenet_2015} validation set is used for clean inputs to the model and serves as the starting point to generate adversarial examples. All experiments are run across 12 A100 GPUs with 40 GB of VRAM and CUDA version 11.1 or greater. Our code is available for download at \url{https://github.com/kyangl/alignment-security}.

\subsection{Average Alignment}
We first investigate how well different classes of alignment---neural alignment, behavioral alignment, and engineering task performance---predict model robustness. For each class, we take the average score across all the benchmarks, giving us a single score for each model in the class. While many works have typically studied average vision alignment overall (i.e., the average of all the benchmarks across all classes), it has been shown that this can overemphasize behavioral alignment at the cost of neural alignment~\cite{ahlert_how_2024}. 

We then assessed each model's robust accuracy against AutoAttack at three different values of epsilon $\epsilon = \{0.001, 0.00196, 0.00392\}$, which corresponds to $\{\frac{0.25}{255}, \frac{0.5}{255}, \frac{1}{255}\}$, respectively. In \autoref{fig:avg_vision_robustness}, we analyze the average score for neural alignment, behavioral alignment, and engineering task performance on the x-axis and the robust accuracy on the y-axis. Each dot represents a model, and the 3 colors correspond to the model's robust accuracy at the three different epsilon values. We perform linear fits for each epsilon value and report the statistical significance. 

We find statistically significant ($p \leq 0.05$) correlations for: the two lowest $\epsilon$ values for neural alignment (explaining up to 9\% of variance), $\epsilon=0.001$ for behavioral alignment (6\% of variance), and at the two lowest $\epsilon$ values for engineering task performance (up to 19\% of variance). Overall, the relatively low $R^2$ values, coupled with the difficulty of getting statistically significant correlations at higher epsilon values, suggest that average alignment scores are, at best, a weak indicator of robust accuracy. This counter-intuitive finding leads us to study further on individual benchmarks. 

\subsection{Individual Benchmarks}
Motivated by the previous experiment where we find that average alignment is weakly correlated with robust accuracy, we hypothesize that averaging scores across different benchmarks may obscure that some individual benchmarks are stronger predictors of robust accuracy than others. Here, certain alignment benchmarks could be more indicative of robust accuracy than others. 

To examine this hypothesis, we collect all models' scores on individual benchmarks for the three classes (neural alignment, behavioral alignment, and engineering task performance) and compute the correlation between each of these scores and robust accuracy at our three different $\epsilon$ values. \autoref{fig:robust_only_correlation} shows a heatmap of the 105 different benchmarks on the x-axis and robust accuracy at three different $\epsilon$ values on the y-axis. In each cell, we report the Spearman correlation coefficient between the selected benchmark score and robust accuracy across models.

From this figure, we find multiple interesting trends. First, we see a wide range of correlations between different benchmarks, confirming our hypothesis that not every current alignment metric is a good indicator of robust accuracy. Additionally, we sometimes see significant changes to the correlation of robust accuracy and a benchmark as the $\epsilon$ value increases (and thus becomes a stronger attack). These changes appear to cluster by class of alignment. Roughly speaking, the neural alignment benchmarks (shown from the first to the second black bar) seem highly dependent on the task, with benchmarks in this category having correlations at both ends of the spectrum. The behavioral benchmarks (shown from the second to third black bars) tend to be, surprisingly, often negatively correlated with robust accuracy at mid and high $\epsilon$ values, and the correlation mostly decreases as $\epsilon$ increases. Finally, engineering task performance (shown from the third black bar to the end of the figure) tends to have more stable (and more positive) correlations as $\epsilon$ increases.

Furthermore, we identified specific benchmarks that are strongly correlated with robust accuracy. Interestingly, many of these benchmarks that exhibit strong positive correlations with robust accuracy even at high epsilon values measure, to some degree, a model's selectivity towards texture information, meaning that \textit{models which are more robust to adversarial examples tend to process texture information more similarly to humans.} Below, we highlight and discuss the benchmarks we found to be most strongly (positive or negative) correlated with robust accuracy. 

In the neural category, we found the correlation between robust accuracy and alignment to be highly dependent on the area of visual processing. At small to medium values of $\epsilon$, correlations are strongest in benchmarks that measure alignment to V4 and IT areas of the visual system. At later values of $\epsilon$, which represents a stronger attack, \texttt{FreemanZiemba2013.V2-pls} which measures responses to naturalistic texture stimuli in V2~\cite{freeman_functional_2013}, was the metric with the strongest positive correlation. This suggests that processing texture more similarly to biological vision systems in later visual areas is even more important than early visual areas to yield robust models. Interestingly, we found that metrics utilizing fMRI data were typically the most negatively correlated.

In the behavioral category, we see many strong positive correlations at low values of $\epsilon$. The sets of benchmarks that stand out the most are the BMD and Baker~\cite{baker_deep_2022} groups, which measure the ability to recognize objects by silhouettes when the shape has been distorted. In \citet{baker_deep_2022}, it was found that humans are less able to recognize objects when shape information is distorted than neural networks are, demonstrating that machine learning models have a certain degree of insensitivity to changes in shape information and thus don't rely as heavily on it as humans do for object recognition. This finding further supports the notion that neural networks may rely more on texture information rather than shape information~\cite{geirhos_imagenet-trained_2019}. The strong positive relationship between these benchmarks and performance on adversarial examples suggests that \textit{models' ability to recognize objects without spatial relationships actually hurts their overall robustness.}

Finally, on engineering tasks, we see strong positive correlations across nearly all benchmarks with low $\epsilon$. However, these correlations decrease as $\epsilon$ increases for most benchmarks. The benchmarks that remain most strongly positively correlated with high $\epsilon$ are both related to bias towards shape over texture information. First is the \texttt{Geirhos2021cueconflict-top1} benchmark~\cite{geirhos_imagenet-trained_2019}, which measures the probability of a model classifying an object using shape information rather than texture via texture-shape conflicted images. The other is the set of benchmarks from~\cite{hermann_origins_2020}: \texttt{Hermann 2020 cue conflict shape \_bias} and \texttt{Hermann2020cueconflictshape\_match}, which similarly measures the probability of a model classifying an object using shape information and the percentage of the times the model classifies according to the shape class, rather than texture or other classes. In all, these results show that when models are able to classify inputs according to their shape information, as humans do, rather than texture information, they will be more robust to adversarial examples.

% we found strong positive correlations in \texttt{FreemanZiemba2013.V1-pls} and \texttt{FreemanZiemba2013.V2-pls} from~\cite{freeman_functional_2013}, which measures neural responses in V1 and V2 to naturalistic texture stimuli. This suggests a link between texture processing in early visual areas and robustness. In the engineering category, two sets of benchmarks stood out as having strong correlations with robust accuracy. First is the \texttt{Geirhos2021cueconflict-top1} benchmark from~\cite{geirhos_imagenet-trained_2019}, which measures the probability of a model classifying an object using shape information rather than texture via  texture-shape conflicted images. A higher score here indicates a stronger shape bias, which appears to be beneficial for robustness. The other is the set of benchmarks from~\cite{hermann_origins_2020}: \texttt{Hermann 2020 cue conflict shape \_bias} and \texttt{Hermann2020cueconflictshape\_match}, which similarly measures the probability of a model classifying an object using shape information and the percentage of the times the model classifies according to the shape class, rather than texture or other classes. This set of benchmarks also show strong positive correlations with model robustness. 

Across all categories of visual alignment, we find that models which exhibit a stronger reliance on shape information and robust processing of texture cues (i.e., more aligned to how humans process information) tend to be more resilient to adversarial examples. These results suggest that increasing alignment towards preferring and processing high-level visual features, particularly textures, in the way biological vision systems do, serves as a fruitful direction for creating more robust and more aligned models.

\subsection{Robustness in Alignment Space}
Building on our finding that certain alignment benchmarks (especially those measuring texture sensitivity) are strongly correlated with adversarial robustness, we next explore whether models with similar alignment profiles also exhibit similar levels of robust accuracy. The goal of this analysis is to see if model similarity on alignment benchmarks is predictive of model performance against adversarial examples, which would support the hypothesis that aligning models under certain metrics results in better robustness.

\begin{figure}[t]
    \centering
    \includegraphics[width=\linewidth]{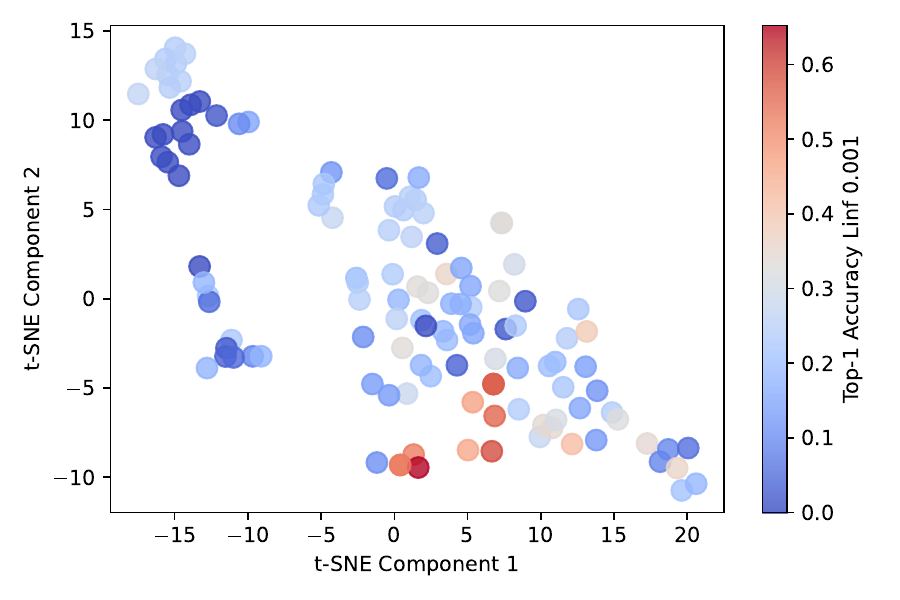}
    \caption{t-SNE plot of principal components from benchmark features for each model colored by the model's robust accuracy.}
    \label{fig:tsne}
    \vspace{0.5cm}
\end{figure}

To compare the similarity of model performance across benchmarks, we utilize t-SNE to reduce the dimensions of the results and project them into a subspace that can be visualized. In \autoref{fig:tsne}, each point represents one of our \nummodels{} models, colored by its robust accuracy under $\ell_\infty$ adversarial perturbations with $\epsilon = 0.001$. We observe that the models are distributed across the embedding space with clustering patterns that are unique to their robustness.

% Analyzing the distribution of non-robust models (blue dots) we notice some clustering of least robust models in the upper left corner, but overall the non-robust models are widely spread across the t-SNE subspace. This demonstrates that vulnerability cannot be attributed to a specific combination of alignment scores but rather that \textit{there are many different ways in which a model can be not robust}. 

Analyzing the non-robust models (blue dots), we find that while some cluster in the upper left, many are scattered broadly throughout the space. This dispersion suggests that poor robustness is not associated with a single alignment profile. Rather, there are \textit{many different ways in which models can fail to defend against adversarial examples}. That is, vulnerability appears to be distributed across a range of misaligned configurations.

% In contrast, the robust models (red dots), we see the opposite behavior of the non-robust models; the models which have the highest degree of robustness to adversarial examples are very well clustered in the subspace and also isolated from the non-robust models. This demonstrates that robust models all share a similar alignment profile, and one that differs distinctly from the non-robust models. This suggests that aligning models in specific ways can lead to higher robustness.

% Examining the non-robust models (blue dots), we find that while some cluster in the upper left, many are scattered broadly throughout the space. This dispersion suggests that poor robustness is not associated with a single alignment profile—rather, there are many different ways in which models can fail to defend against adversarial examples. That is, vulnerability appears to be distributed across a range of misaligned configurations.

In contrast, the robust models (red dots) form a tight, well-separated cluster. These models not only share a similar alignment profile, but their separation from the rest of the models \textit{suggests that robust behavior is tied to a specific region of alignment space}. This strongly supports the view that certain alignment characteristics are predictive of robustness.

These findings offer a new perspective on adversarial robustness; rather than unique quirks of the model resulting in vulnerability to adversarial examples, there are specific, isolated properties that allow models to more robustly process inputs. In other words, \textit{it's not that specific misalignment will lead to model vulnerability, but rather specific alignment will lead to robustness}. Furthermore, these properties can be measured in alignment metrics, and thus optimized for in order to build more robust models.

% cluster of low accuracy and high accuracy. the fact that they cluster, model performance on individual benchmarks are highly predictive of robust accuracy. Demonstrating that increasing alignment should also increase alignment. 

\section{Related Work}\label{sec:related}
There has been substantial progress in bridging the representational differences between humans and machine learning models over the last few years. \citet{geirhos_partial_2021} show that many of the high-performance models match or exceed human feedforward performance on OOD datasets. Several works suggest that more human-aligned model architectures may also be more robust. \citet{dapello_simulating_2020} design the biologically inspired VOne block to simulate V1 processing, improving robustness to adversarial and common corruptions. \citet{li_learning_2019} regularize models based on neural recordings from mice to increase robustness and alignment. 

Furthermore, \citet{subramanian_spatial-frequency_2023} show that differences in spatial frequency processing explain both shape bias and adversarial vulnerability. Models with higher human alignment show improved performance on datasets like ImageNet-A~\cite{sucholutsky_alignment_2023}. Other works highlight that models tend to rely on texture over shape~\cite{geirhos_imagenet-trained_2019, hermann_origins_2020}, a bias that increases susceptibility to natural adversarial samples~\cite{hoak_err_2025, hoak_explorations_2024}.

Our work expands on these findings with a large-scale empirical study across diverse architectures and benchmarks. Unlike prior work that focuses on specific models or alignment methods, we systematically analyze how different forms of alignment relate to adversarial robustness, discovering meaningful connections between model perception and security.

% \section{Introduction}

% The European Conference of Artificial Intelligence (ECAI) is the leading 
% discipline-wide conference on AI in Europe. Its history goes back all 
% the way to the Summer Conference on Artificial Intelligence and 
% Simulation of Behaviour held in July 1974 in Brighton. Nowadays, ECAI is 
% organised annually under the auspices of the European Association for 
% Artificial Intelligence (EurAI, see Figure~\ref{fig:eurai}).

% \begin{figure}[h]
% \centering
% \includegraphics[width=2.5cm]{eurai}
% \caption{Logo of the European Association for Artificial Intelligence.}
% \label{fig:eurai}
% \end{figure}

% Your paper should be typeset in \LaTeX, using the ECAI class file 
% provided (\texttt{ecai.cls}). Please do not modify the class file or 
% any of the layout parameters.

% For instructions on how to submit your work to ECAI and on matters such 
% as page limits or referring to supplementary material, please consult 
% the Call for Papers of the next edition of the conference. Keep in mind
% that you must use the \texttt{doubleblind} option for submission. 

%%%%%%%%%%%%%%%%%%%%%%%%%%%%%%%%%%%%%%%%%%%%%%%%%%%%%%%%%%%%%%%%%%%%%%%%
\section{Conclusions}
% In this work, we found that that while robustness was only weakly correlated with \textit{average} vision alignment, \textit{specific} alignment benchmarks were strongly predictive of adversarial robustness. Additionally, we observed that 1) robustness is well-structured in the alignment space, and 2) certain individual benchmarks serve as strong indicators of robust accuracy, particularly those that assess a model's preference for texture information over shape. From this, we hope to encourage future work to leverage insights found in both areas to build more secure and aligned vision systems.
In this work, we investigated the relationship between representational alignment and adversarial robustness in vision models. While robustness showed only a weak correlation with \textit{average} alignment scores, we found that \textit{specific} alignment benchmarks, particularly those measuring texture selectivity, were strongly predictive of robust accuracy. Additionally, we observed that robustness is well-structured in alignment space: models with similar alignment profiles tend to exhibit similar levels of robustness, and highly robust models occupy a distinct region of this space. These findings suggest that not all alignment is equally beneficial for security and that targeted alignment along specific dimensions is a promising strategy for improving robustness. We hope this work motivates future research at the intersection of alignment and security, with the goal of developing vision systems that are both perceptually grounded and resilient to adversarial threats.

%%%%%%%%%%%%%%%%%%%%%%%%%%%%%%%%%%%%%%%%%%%%%%%%%%%%%%%%%%%%%%%%%%%%%%%%

%%%%%%%%%%%%%%%%%%%%%%%%%%%%%%%%%%%%%%%%%%%%%%%%%%%%%%%%%%%%%%%%%%%%%%%%

%%% Use this environment to include acknowledgements (optional).
%%% This will be omitted in doubleblind mode.

\begin{ack}
This material is based upon work supported by, or in part by, the National Science Foundation under Grant No. CNS 2343611, and by the Combat Capabilities Development Command Army Research Office under Grant No. W911NF-21-1-0317 (ARO MURI). Any opinions, findings, and conclusions or recommendations expressed in this publication are those of the author(s) and do not necessarily reflect the views of the National Science Foundation, the U.S. Government, or the Department of Defense. The U.S. Government is authorized to reproduce and distribute reprints for government purposes notwithstanding any copyright notation hereon.
\end{ack}

%%%%%%%%%%%%%%%%%%%%%%%%%%%%%%%%%%%%%%%%%%%%%%%%%%%%%%%%%%%%%%%%%%%%%%%%

%%% Use this command to include your bibliography file.
\bibliography{references.bib}

\begin{thebibliography}{39}
\providecommand{\natexlab}[1]{#1}
\providecommand{\url}[1]{\texttt{#1}}
\expandafter\ifx\csname urlstyle\endcsname\relax
  \providecommand{\doi}[1]{doi: #1}\else
  \providecommand{\doi}{doi: \begingroup \urlstyle{rm}\Url}\fi

\bibitem[Ahlert et~al.(2024)Ahlert, Klein, Wichmann, and Geirhos]{ahlert_how_2024}
J.~Ahlert, T.~Klein, F.~Wichmann, and R.~Geirhos.
\newblock How {Aligned} are {Different} {Alignment} {Metrics}?, July 2024.
\newblock URL \url{http://arxiv.org/abs/2407.07530}.

\bibitem[Athalye et~al.(2018)Athalye, Carlini, and Wagner]{athalye_obfuscated_2018}
A.~Athalye, N.~Carlini, and D.~Wagner.
\newblock Obfuscated {Gradients} {Give} a {False} {Sense} of {Security}: {Circumventing} {Defenses} to {Adversarial} {Examples}.
\newblock In \emph{Proceedings of the 35th ICML}. PMLR, July 2018.
\newblock URL \url{https://proceedings.mlr.press/v80/athalye18a.html}.

\bibitem[Baker and Elder(2022)]{baker_deep_2022}
N.~Baker and J.~H. Elder.
\newblock Deep learning models fail to capture the configural nature of human shape perception.
\newblock \emph{iScience}, 25\penalty0 (9), Sept. 2022.
\newblock ISSN 2589-0042.
\newblock URL \url{https://www.sciencedirect.com/science/article/pii/S2589004222011853}.

\bibitem[Beyer et~al.(2022)Beyer, Zhai, Royer, Markeeva, Anil, and Kolesnikov]{beyer_knowledge_2022}
L.~Beyer, X.~Zhai, A.~Royer, L.~Markeeva, R.~Anil, and A.~Kolesnikov.
\newblock Knowledge {Distillation}: {A} {Good} {Teacher} {Is} {Patient} and {Consistent}.
\newblock pages 10925--10934, 2022.
\newblock URL \url{https://openaccess.thecvf.com/content/CVPR2022/html/Beyer_Knowledge_Distillation_A_Good_Teacher_Is_Patient_and_Consistent_CVPR_2022_paper.html}.

\bibitem[Carlini and Wagner(2017)]{carlini_towards_2017}
N.~Carlini and D.~Wagner.
\newblock Towards {Evaluating} the {Robustness} of {Neural} {Networks}, Mar. 2017.
\newblock URL \url{http://arxiv.org/abs/1608.04644}.

\bibitem[Carlini et~al.(2019)Carlini, Athalye, Papernot, Brendel, Rauber, Tsipras, Goodfellow, Madry, and Kurakin]{carlini_evaluating_2019}
N.~Carlini, A.~Athalye, N.~Papernot, W.~Brendel, J.~Rauber, D.~Tsipras, I.~Goodfellow, A.~Madry, and A.~Kurakin.
\newblock On {Evaluating} {Adversarial} {Robustness}, Feb. 2019.
\newblock URL \url{http://arxiv.org/abs/1902.06705}.

\bibitem[Cheng et~al.(2024)Cheng, Rodriguez, Chen, Kar, Watanabe, and Serre]{cheng_rtify_2024}
Y.-A. Cheng, I.~F. Rodriguez, S.~Chen, K.~Kar, T.~Watanabe, and T.~Serre.
\newblock {RTify}: {Aligning} {Deep} {Neural} {Networks} with {Human} {Behavioral} {Decisions}, Dec. 2024.
\newblock URL \url{http://arxiv.org/abs/2411.03630}.

\bibitem[Croce and Hein(2020)]{croce_reliable_2020}
F.~Croce and M.~Hein.
\newblock Reliable evaluation of adversarial robustness with an ensemble of diverse parameter-free attacks.
\newblock In \emph{Proceedings of the 37th {International} {Conference} on {Machine} {Learning}}. PMLR, Nov. 2020.
\newblock URL \url{https://proceedings.mlr.press/v119/croce20b.html}.

\bibitem[Croce et~al.(2021)Croce, Andriushchenko, Sehwag, Debenedetti, Flammarion, Chiang, Mittal, and Hein]{croce_robustbench_2021}
F.~Croce, M.~Andriushchenko, V.~Sehwag, E.~Debenedetti, N.~Flammarion, M.~Chiang, P.~Mittal, and M.~Hein.
\newblock {RobustBench}: a standardized adversarial robustness benchmark.
\newblock In \emph{{NeurIPS}}. arXiv, Oct. 2021.
\newblock URL \url{http://arxiv.org/abs/2010.09670}.

\bibitem[Dapello et~al.(2020)Dapello, Marques, Schrimpf, Geiger, Cox, and DiCarlo]{dapello_simulating_2020}
J.~Dapello, T.~Marques, M.~Schrimpf, F.~Geiger, D.~Cox, and J.~J. DiCarlo.
\newblock Simulating a {Primary} {Visual} {Cortex} at the {Front} of {CNNs} {Improves} {Robustness} to {Image} {Perturbations}.
\newblock In \emph{Advances in {Neural} {Information} {Processing} {Systems}}, volume~33, pages 13073--13087. Curran Associates, Inc., 2020.
\newblock URL \url{https://proceedings.neurips.cc/paper/2020/hash/98b17f068d5d9b7668e19fb8ae470841-Abstract.html}.

\bibitem[Dosovitskiy et~al.(2021)Dosovitskiy, Beyer, Kolesnikov, Weissenborn, Zhai, Unterthiner, Dehghani, Minderer, Heigold, Gelly, Uszkoreit, and Houlsby]{dosovitskiy_image_2021}
A.~Dosovitskiy, L.~Beyer, A.~Kolesnikov, D.~Weissenborn, X.~Zhai, T.~Unterthiner, M.~Dehghani, M.~Minderer, G.~Heigold, S.~Gelly, J.~Uszkoreit, and N.~Houlsby.
\newblock An {Image} is {Worth} 16x16 {Words}: {Transformers} for {Image} {Recognition} at {Scale}, June 2021.
\newblock URL \url{http://arxiv.org/abs/2010.11929}.

\bibitem[Freeman et~al.(2013)Freeman, Ziemba, Heeger, Simoncelli, and Movshon]{freeman_functional_2013}
J.~Freeman, C.~M. Ziemba, D.~J. Heeger, E.~P. Simoncelli, and J.~A. Movshon.
\newblock A functional and perceptual signature of the second visual area in primates.
\newblock \emph{Nature Neuroscience}, 16\penalty0 (7):\penalty0 974--981, July 2013.
\newblock ISSN 1546-1726.
\newblock URL \url{https://www.nature.com/articles/nn.3402}.

\bibitem[Geirhos et~al.(2019)Geirhos, Rubisch, Michaelis, Bethge, Wichmann, and Brendel]{geirhos_imagenet-trained_2019}
R.~Geirhos, P.~Rubisch, C.~Michaelis, M.~Bethge, F.~A. Wichmann, and W.~Brendel.
\newblock {ImageNet}-trained {CNNs} are biased towards texture; increasing shape bias improves accuracy and robustness.
\newblock In \emph{{ICLR}}, Jan. 2019.
\newblock URL \url{http://arxiv.org/abs/1811.12231}.

\bibitem[Geirhos et~al.(2020)Geirhos, Temme, Rauber, Schütt, Bethge, and Wichmann]{geirhos_generalisation_2020}
R.~Geirhos, C.~R.~M. Temme, J.~Rauber, H.~H. Schütt, M.~Bethge, and F.~A. Wichmann.
\newblock Generalisation in humans and deep neural networks.
\newblock In \emph{{NeurIPS} 2018}, 2020.
\newblock URL \url{http://arxiv.org/abs/1808.08750}.

\bibitem[Geirhos et~al.(2021)Geirhos, Narayanappa, Mitzkus, Thieringer, Bethge, Wichmann, and Brendel]{geirhos_partial_2021}
R.~Geirhos, K.~Narayanappa, B.~Mitzkus, T.~Thieringer, M.~Bethge, F.~A. Wichmann, and W.~Brendel.
\newblock Partial success in closing the gap between human and machine vision.
\newblock In \emph{35th {Conference} on NeurIPS}. NeurIPS, Oct. 2021.
\newblock URL \url{http://arxiv.org/abs/2106.07411}.

\bibitem[Goodfellow et~al.(2015)Goodfellow, Shlens, and Szegedy]{goodfellow_explaining_2015}
I.~J. Goodfellow, J.~Shlens, and C.~Szegedy.
\newblock Explaining and {Harnessing} {Adversarial} {Examples}, Mar. 2015.
\newblock URL \url{http://arxiv.org/abs/1412.6572}.

\bibitem[He et~al.(2015)He, Zhang, Ren, and Sun]{he_deep_2015}
K.~He, X.~Zhang, S.~Ren, and J.~Sun.
\newblock Deep {Residual} {Learning} for {Image} {Recognition}.
\newblock In \emph{{CVPR} 2016}, 2015.
\newblock URL \url{http://arxiv.org/abs/1512.03385}.

\bibitem[Hermann et~al.(2020)Hermann, Chen, and Kornblith]{hermann_origins_2020}
K.~L. Hermann, T.~Chen, and S.~Kornblith.
\newblock The {Origins} and {Prevalence} of {Texture} {Bias} in {Convolutional} {Neural} {Networks}.
\newblock In \emph{{NeurIPS} 2020}, Nov. 2020.
\newblock URL \url{http://arxiv.org/abs/1911.09071}.

\bibitem[Hoak and McDaniel(2024)]{hoak_explorations_2024}
B.~Hoak and P.~McDaniel.
\newblock Explorations in {Texture} {Learning}.
\newblock In \emph{{ICLR} 2024, {Tiny} {Papers} {Track}}, 2024.
\newblock URL \url{http://arxiv.org/abs/2403.09543}.

\bibitem[Hoak et~al.(2025)Hoak, Sheatsley, and McDaniel]{hoak_err_2025}
B.~Hoak, R.~Sheatsley, and P.~McDaniel.
\newblock Err on the {Side} of {Texture}: {Texture} {Bias} on {Real} {Data}.
\newblock In \emph{2025 {IEEE} {Conference} on {Secure} and {Trustworthy} {Machine} {Learning} ({SaTML})}, pages 661--680. IEEE Computer Society, Apr. 2025.
\newblock URL \url{https://www.computer.org/csdl/proceedings-article/satml/2025/171100a661/26Vnph1kKxW}.

\bibitem[Kim(2021)]{kim_torchattacks_2021}
H.~Kim.
\newblock Torchattacks: {A} {PyTorch} {Repository} for {Adversarial} {Attacks}, Feb. 2021.
\newblock URL \url{http://arxiv.org/abs/2010.01950}.

\bibitem[Kornblith et~al.(2019)Kornblith, Norouzi, Lee, and Hinton]{kornblith_similarity_2019}
S.~Kornblith, M.~Norouzi, H.~Lee, and G.~Hinton.
\newblock Similarity of {Neural} {Network} {Representations} {Revisited}, July 2019.
\newblock URL \url{http://arxiv.org/abs/1905.00414}.

\bibitem[Kriegeskorte et~al.(2008)Kriegeskorte, Mur, and Bandettini]{kriegeskorte_representational_2008}
N.~Kriegeskorte, M.~Mur, and P.~A. Bandettini.
\newblock Representational similarity analysis - connecting the branches of systems neuroscience.
\newblock \emph{Frontiers in Systems Neuroscience}, 2, Nov. 2008.
\newblock ISSN 1662-5137.
\newblock URL \url{https://www.frontiersin.org/journals/systems-neuroscience/articles/10.3389/neuro.06.004.2008/full}.

\bibitem[Krizhevsky et~al.(2017)Krizhevsky, Sutskever, and Hinton]{krizhevsky_imagenet_2017}
A.~Krizhevsky, I.~Sutskever, and G.~E. Hinton.
\newblock {ImageNet} classification with deep convolutional neural networks.
\newblock \emph{Communications of the ACM}, 60\penalty0 (6), May 2017.
\newblock URL \url{https://dl.acm.org/doi/10.1145/3065386}.

\bibitem[Li et~al.(2019)Li, Brendel, Walker, Cobos, Muhammad, Reimer, Bethge, Sinz, Pitkow, and Tolias]{li_learning_2019}
Z.~Li, W.~Brendel, E.~Walker, E.~Cobos, T.~Muhammad, J.~Reimer, M.~Bethge, F.~Sinz, Z.~Pitkow, and A.~Tolias.
\newblock Learning from brains how to regularize machines.
\newblock In \emph{Advances in {Neural} {Information} {Processing} {Systems}}, volume~32. Curran Associates, Inc., 2019.
\newblock URL \url{https://proceedings.neurips.cc/paper/2019/hash/70117ee3c0b15a2950f1e82a215e812b-Abstract.html}.

\bibitem[Liu et~al.(2022)Liu, Mao, Wu, Feichtenhofer, Darrell, and Xie]{liu_convnet_2022}
Z.~Liu, H.~Mao, C.-Y. Wu, C.~Feichtenhofer, T.~Darrell, and S.~Xie.
\newblock A {ConvNet} for the 2020s.
\newblock In \emph{Proceedings of the {IEEE}/{CVF} {Conference} on {Computer} {Vision} and {Pattern} {Recognition}}, pages 11976--11986, 2022.
\newblock URL \url{https://openaccess.thecvf.com/content/CVPR2022/html/Liu_A_ConvNet_for_the_2020s_CVPR_2022_paper.html}.

\bibitem[Madry et~al.(2019)Madry, Makelov, Schmidt, Tsipras, and Vladu]{madry_towards_2019}
A.~Madry, A.~Makelov, L.~Schmidt, D.~Tsipras, and A.~Vladu.
\newblock Towards {Deep} {Learning} {Models} {Resistant} to {Adversarial} {Attacks}, Sept. 2019.
\newblock URL \url{http://arxiv.org/abs/1706.06083}.

\bibitem[Muttenthaler et~al.(2023)Muttenthaler, Linhardt, Dippel, Vandermeulen, Hermann, Lampinen, and Kornblith]{muttenthaler_improving_2023}
L.~Muttenthaler, L.~Linhardt, J.~Dippel, R.~A. Vandermeulen, K.~Hermann, A.~K. Lampinen, and S.~Kornblith.
\newblock Improving neural network representations using human similarity judgments, Sept. 2023.
\newblock URL \url{http://arxiv.org/abs/2306.04507}.

\bibitem[O'Shea and Nash(2015)]{oshea_introduction_2015}
K.~O'Shea and R.~Nash.
\newblock An {Introduction} to {Convolutional} {Neural} {Networks}, Dec. 2015.
\newblock URL \url{http://arxiv.org/abs/1511.08458}.

\bibitem[Radford et~al.(2021)Radford, Kim, Hallacy, Ramesh, Goh, Agarwal, Sastry, Askell, Mishkin, Clark, Krueger, and Sutskever]{radford_learning_2021}
A.~Radford, J.~W. Kim, C.~Hallacy, A.~Ramesh, G.~Goh, S.~Agarwal, G.~Sastry, A.~Askell, P.~Mishkin, J.~Clark, G.~Krueger, and I.~Sutskever.
\newblock Learning {Transferable} {Visual} {Models} {From} {Natural} {Language} {Supervision}.
\newblock In \emph{Proceedings of the 38th ICML}. PMLR, July 2021.
\newblock URL \url{https://proceedings.mlr.press/v139/radford21a.html}.

\bibitem[Russakovsky et~al.(2015)Russakovsky, Deng, Su, Krause, Satheesh, Ma, Huang, Karpathy, Khosla, Bernstein, Berg, and Fei-Fei]{russakovsky_imagenet_2015}
O.~Russakovsky, J.~Deng, H.~Su, J.~Krause, S.~Satheesh, S.~Ma, Z.~Huang, A.~Karpathy, A.~Khosla, M.~Bernstein, A.~C. Berg, and L.~Fei-Fei.
\newblock {ImageNet} {Large} {Scale} {Visual} {Recognition} {Challenge}.
\newblock In \emph{{IJCV} 2015}. arXiv, 2015.
\newblock URL \url{http://arxiv.org/abs/1409.0575}.

\bibitem[Schrimpf et~al.(2018)Schrimpf, Kubilius, Hong, Majaj, Rajalingham, Issa, Kar, Bashivan, Prescott-Roy, Geiger, Schmidt, Yamins, and DiCarlo]{schrimpf_brain-score_2018}
M.~Schrimpf, J.~Kubilius, H.~Hong, N.~J. Majaj, R.~Rajalingham, E.~B. Issa, K.~Kar, P.~Bashivan, J.~Prescott-Roy, F.~Geiger, K.~Schmidt, D.~L.~K. Yamins, and J.~J. DiCarlo.
\newblock Brain-{Score}: {Which} {Artificial} {Neural} {Network} for {Object} {Recognition} is most {Brain}-{Like}?, Sept. 2018.
\newblock URL \url{http://biorxiv.org/lookup/doi/10.1101/407007}.

\bibitem[Sheatsley et~al.(2023)Sheatsley, Hoak, Pauley, and McDaniel]{sheatsley_space_2023}
R.~Sheatsley, B.~Hoak, E.~Pauley, and P.~McDaniel.
\newblock The {Space} of {Adversarial} {Strategies}.
\newblock In \emph{32nd {USENIX} {Security} {Symposium}}, 2023.
\newblock URL \url{https://www.usenix.org/conference/usenixsecurity23/presentation/sheatsley}.

\bibitem[Simonyan et~al.(2014)Simonyan, Vedaldi, and Zisserman]{simonyan_deep_2014}
K.~Simonyan, A.~Vedaldi, and A.~Zisserman.
\newblock Deep {Inside} {Convolutional} {Networks}: {Visualising} {Image} {Classification} {Models} and {Saliency} {Maps}, Apr. 2014.
\newblock URL \url{http://arxiv.org/abs/1312.6034}.

\bibitem[Subramanian et~al.(2023)Subramanian, Sizikova, Majaj, and Pelli]{subramanian_spatial-frequency_2023}
A.~Subramanian, E.~Sizikova, N.~J. Majaj, and D.~G. Pelli.
\newblock Spatial-frequency channels, shape bias, and adversarial robustness.
\newblock In \emph{Conference on {Neural} {Information} {Processing} {Systems}}. NeurIPS, 2023.

\bibitem[Sucholutsky and Griffiths(2023)]{sucholutsky_alignment_2023}
I.~Sucholutsky and T.~L. Griffiths.
\newblock Alignment with human representations supports robust few-shot learning.
\newblock In \emph{Advances in {Neural} {Information} {Processing} {Systems}}, volume~36, pages 73464--73479, Oct. 2023.
\newblock URL \url{https://proceedings.neurips.cc/paper_files/paper/2023/hash/e8ddc03b001d4c4b44b29bc1167e7fdd-Abstract-Conference.html}.

\bibitem[Sucholutsky et~al.(2024)Sucholutsky, Muttenthaler, Weller, Peng, Bobu, Kim, Love, Cueva, Grant, Groen, Achterberg, Tenenbaum, Collins, Hermann, Oktar, Greff, Hebart, Cloos, Kriegeskorte, Jacoby, Zhang, Marjieh, Geirhos, Chen, Kornblith, Rane, Konkle, O'Connell, Unterthiner, Lampinen, Müller, Toneva, and Griffiths]{sucholutsky_getting_2024}
I.~Sucholutsky, L.~Muttenthaler, A.~Weller, A.~Peng, A.~Bobu, B.~Kim, B.~C. Love, C.~J. Cueva, E.~Grant, I.~Groen, J.~Achterberg, J.~B. Tenenbaum, K.~M. Collins, K.~L. Hermann, K.~Oktar, K.~Greff, M.~N. Hebart, N.~Cloos, N.~Kriegeskorte, N.~Jacoby, Q.~Zhang, R.~Marjieh, R.~Geirhos, S.~Chen, S.~Kornblith, S.~Rane, T.~Konkle, T.~P. O'Connell, T.~Unterthiner, A.~K. Lampinen, K.-R. Müller, M.~Toneva, and T.~L. Griffiths.
\newblock Getting aligned on representational alignment, Nov. 2024.
\newblock URL \url{http://arxiv.org/abs/2310.13018}.
\newblock arXiv:2310.13018 [q-bio].

\bibitem[Touvron et~al.(2021)Touvron, Cord, Douze, Massa, Sablayrolles, and Jegou]{touvron_training_2021}
H.~Touvron, M.~Cord, M.~Douze, F.~Massa, A.~Sablayrolles, and H.~Jegou.
\newblock Training data-efficient image transformers \& distillation through attention.
\newblock In \emph{Proceedings of the 38th {International} {Conference} on {Machine} {Learning}}. PMLR, July 2021.
\newblock URL \url{https://proceedings.mlr.press/v139/touvron21a.html}.

\bibitem[Yamins et~al.(2013)Yamins, Hong, Cadieu, and DiCarlo]{yamins_hierarchical_2013}
D.~L. Yamins, H.~Hong, C.~Cadieu, and J.~J. DiCarlo.
\newblock Hierarchical {Modular} {Optimization} of {Convolutional} {Networks} {Achieves} {Representations} {Similar} to {Macaque} {IT} and {Human} {Ventral} {Stream}.
\newblock In \emph{Advances in {Neural} {Information} {Processing} {Systems}}, volume~26. Curran Associates, Inc., 2013.
\newblock URL \url{https://papers.nips.cc/paper_files/paper/2013/hash/9a1756fd0c741126d7bbd4b692ccbd91-Abstract.html}.

\end{thebibliography}

\end{document}